\documentclass[10pt,twocolumn,letterpaper]{article}

\usepackage{iccv}
\usepackage{times}
\usepackage{epsfig}
\usepackage{graphicx}
\usepackage{amsmath}
\usepackage{amssymb}
\usepackage{arydshln}
\usepackage[pagebackref=true,breaklinks=true,letterpaper=true,colorlinks,bookmarks=false]{hyperref}
\usepackage{fancyhdr}
\usepackage[paper=a4paper,headsep=0.2cm, headheight=0.75cm,top=1.8cm]{geometry} 

\iccvfinalcopy

\def\iccvPaperID{8}

\fancypagestyle{empty}{%
  \fancyhead{}
    \fancyhfoffset[L]{2cm} 
    \fancyhfoffset[R]{2cm}
    \fancyhead[C]{Published at ICCV 2023 - BioImage Computing Workshop}
}

\begin{document}

\title{Towards Hierarchical Regional Transformer-based Multiple Instance Learning}

\author{Josef Cersovsky\\
Bayer AG\\
Berlin, Germany\\
{\tt\small josef.cersovsky@bayer.com}
\and
Sadegh Mohammadi\\
Bayer AG\\
Berlin, Germany\\
{\tt\small sadegh.mohammadi@bayer.com}
\and
Dagmar Kainmueller\\
Max-Delbrueck-Center for Molecular Medicine\\
Berlin, Germany\\
{\tt\small dagmar.kainmueller@mdc-berlin.de}
\and
Johannes Hoehne\\
Bayer AG\\
Berlin, Germany\\
{\tt\small johannes.hoehne@bayer.com}
}

\maketitle

\ificcvfinal\thispagestyle{empty}\fi

\begin{abstract}
   The classification of gigapixel histopathology images with deep multiple instance learning models has become a critical task in digital pathology and precision medicine. In this work, we propose a Transformer-based multiple instance learning approach that replaces the traditional learned attention mechanism with a regional, Vision Transformer inspired self-attention mechanism. We present a method that fuses regional patch information to derive slide-level predictions and show how this regional aggregation can be stacked to hierarchically process features on different distance levels. To increase predictive accuracy, especially for datasets with small, local morphological features, we introduce a method to focus the image processing on high attention regions during inference. Our approach is able to significantly improve performance over the baseline on two histopathology datasets and points towards promising directions for further research.
\end{abstract}

\pagestyle{fancy}
\fancyhead{}
\fancyhfoffset[L]{2cm} 
\fancyhfoffset[R]{2cm}
\fancyhead[C]{Published at ICCV 2023 - BioImage Computing Workshop}

\section{Introduction}

\begin{figure*}
    \centering
    \includegraphics[width=1\textwidth]{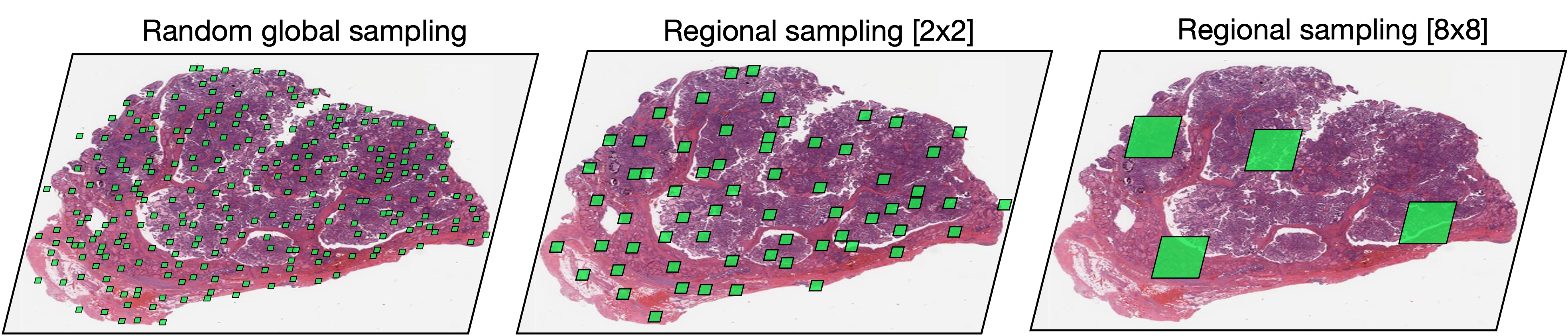}
    \caption{Overview of strategies (patch/region sizes are increased for better visualization). Every depicted approach samples a total of 256 patches. (Left) Standard random global sampling, where all patches are selected randomly across the image. Captures the largest amount of variety, but does not retain any information about regionality. (Center) Regional $2\times2$ sampling. This approach does pertain regional information by selecting regions, but still covers a relatively large spread of image information. (Right) $8\times8$ sampling, used for multi-level aggregation. The large region-size is able to densely capture large morphological structures, but does thereby lead to a less representative sample of the full slide.}
    \label{fig:sampling_strategies}
\end{figure*}

Applying conventional computer vision models to gigapixel images proves to be difficult because of the large computational capacity required. The multiple instance learning (MIL) paradigm \cite{dietterich1997solving} allows treating an image as a non-exhaustive bag of sampled image patches, which retains the image label. A MIL-model enables generating image level predictions by either processing the patches individually or as an aggregated bag representation, thereby drastically reducing input dimensionality. A variety of MIL-based approaches working on instance-level, bag-level or a combination of both have been applied to a range of use cases \cite{amores2013multiple,carbonneau2018multiple,gadermayr22multiple}. This holds especially true in the medical domain \cite{couture2022deep,van2021deep}, where there exists an abundance of high-resolution imaging data. MIL is commonly applied to such images by randomly sampling a bag of image patches to represent the image. However, the incomplete, sparse and often noisy nature of random patch-sampling remains a challenging hurdle and a limiting factor to predictive performance \cite{carbonneau2018multiple, gadermayr22multiple}.

Most current MIL approaches employ some form of attention-based bag aggregation \cite{ilse2018attention}. They generally follow three main steps: First, the image patches are encoded into a latent space by a patch-embedder, usually a convolutional neural network (CNN). Secondly, the bag of patch-embeddings is aggregated into a single bag-level representation using a learned attention mechanism. Finally, a linear classifier derives a prediction from the bag embedding. There are three main advantages to this attention-based architecture: it greatly reduces the dimensionality by embedding patches and aggregating bags, it is invariant to the size and order of the bags and it provides high interpretability through the generated attention weights. Research has shown this approach to generalize across a variety of use cases in digital medicine \cite{gadermayr22multiple,hohne2021detecting,van2021deep}.

Subsequent methods extend and improve attention-powered models applied to MIL \cite{li2019patch}. In recent research, Transformers \cite{vaswani2017attention} have become the state-of-the-art for attention-based learning across many domains, including the Vision Transformer (ViT) for computer vision \cite{dosovitskiy2020image}. Input embeddings are passed through self-attention modules, where they are aggregated using a learned classification token.  Transformer-based models enable a highly effective application of attention to MIL problems \cite{tian2022contrastive,shao2021transmil,myronenko2021accounting}. The RegionViT \cite{chen2021regionvit} separates patches into regions and facilitates global information exchange through a regional token. Chen et al. \cite{chen2022scaling} even demonstrate a method that allows the processing of complete whole slide images by aggregating increasingly large regions in multiple steps through a learned classification token. However, ViT-based models can be computationally costly. They therefore usually require expensive contrastive pre-training and frozen weights for most layers. By sampling patches regionally, we reduce the required memory footprint drastically, allowing us to apply Transformers for end-to-end model training.

Various approaches also incorporate a variant of local attention based patch aggregation, reducing computational complexity and allowing the model to exploit information about patch locality \cite{chen2022scaling,wang2020hierarchical}. Konstantinov et al. \cite{konstantinov2022multi} propose a mechanism that enriches patch embeddings by adding an aggregated representation of the patch neighborhood based on patch similarity. However, using simple dot-product attention to aggregate patches can lose valuable information during the aggregation process. We address this employing a Vision Transformer-like encoder operating on embedding-level for bag aggregation. Because using the Transformer to embed pixel-level information directly, as done in the regular ViT, leads to high computational demands, we use a CNN to derive patch embeddings.

Research has further shown that processing image patches in multiple stages can improve predictive performance \cite{fuster2021nested,tu2021hamil}. Myronenko et al. \cite{myronenko2021accounting} aggregate features on different scales by downscaling the inputs in several steps using a CNN backbone. We propose a novel method of multi-level hierarchical aggregation that does not require downsampling of patches but instead enables processing of variable region sizes by iteratively aggregating increasingly large regions.

There exist several methods of pre-selecting relevant patches to guide model predictions. However, they usually require specialized architectures or additional steps during training \cite{li2021dual,li2021multi,yao2020whole,thandiackal2022differentiable}. To leverage patch-level weights without changes to the model architecture, we instead propose a clustering based high-attention patch selection method that can be used during inference to significantly improve model performance.

\subsection*{Contribution}

This work aims to fuse recent developments in Transformers and MIL, by contributing a novel, regional, hierarchical Transformer architecture for efficient model training on gigapixel images. We demonstrate that our approach can leverage regional patch aggregation and the hierarchically fusion of local information on multiple levels to significantly improve over a naive global patch aggregation baseline \cite{ilse2018attention}. Furthermore, we propose an easily transferable model-agnostic method for processing high-attention patches during inference, demonstrating large performance improvements on a dataset with small regions of interest. Our results showcase the potential of the proposed methodological ideas and build the groundwork for further experimentation and benchmarking.

\section{Method}

\begin{figure*}
    \centering
    \includegraphics[width=1\textwidth]{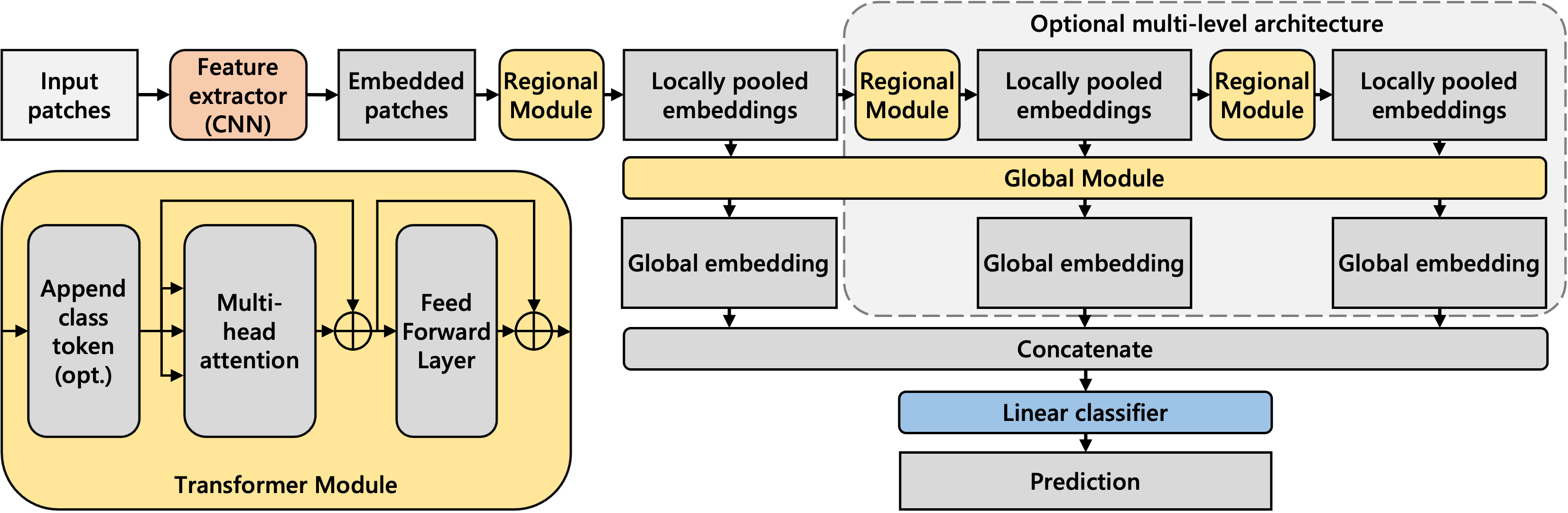}
    \caption{Proposed model architecture. A feature extraction module, usually a pretrained CNN, embeds each patch separately. Embeddings are then pooled locally by a Transformer using learnable class tokens. If multiple transformer modules are stacked, the class token is only appended in the last module. The weighted sum with regard to the class token can be considered as an embedding for the whole region. This local pooling can optionally be repeated multiple times to aggregate larger image regions. Pooled regional aggregations are finally aggregated on a slide level using a global Transformer module to generate slide-level representations. If multiple pooling-levels are used, each level's tokens are aggregated into a separate global representation which are then concatenated. The full slide representation is fed into a linear classifier. In our experiments, weights between the regional modules are shared, while the global module uses separate weights.}
    \label{fig:architecture}
\end{figure*}

\subsection{Multiple Instance Learning}

In multiple instance learning, there exist bags of unordered instances $\hat{X} = \{x_1, ..., x_k\}$. When applied to computer vision, each instance $x_k$ is a subset of an input image $X$, usually corresponding to a comparatively small patch of $X$. Each bag is given the label $Y$ associated with $X$. The labels $y_k$ of individual instances are unknown during training. MIL models aim to determine $Y$ by inferring the instance labels $y_k$ or by aggregating the instances into a bag-level representation $\mathbf{Z}$ from which $Y$ can be deduced. In attention-based MIL, this bag representation $\mathbf{Z}$ is based on a weighted sum of all instances inside the bag,

\begin{equation}
    \label{eqn:bag_aggregation}
    \mathbf{Z} = \sum_{k=1}^{K}a_kx_k,
\end{equation}

where the attention is based on the dot-product similarity of $x_k$ in regard to a learnable parameter $\mathbf{C}$ with the same dimensionality as $x_k$:

\begin{equation}
    \label{eqn:attention_weights}
    a_k = \frac{\exp(\mathbf{C}x_k^\top)}{\sum_{i=1}^K\exp(\mathbf{C}x_i^\top)}.
\end{equation}

We use attention-based MIL applied to patches sampled randomly from whole slide images (WSI) as done in \cite{ilse2018attention} as our baseline. Each patch is a separate instance, which is represented by an embedding derived using a CNN. Note that we benchmark our method against a common, but conceptually simple baseline on two binary classification tasks. This is done on purpose, due to the paper’s focus on demonstrating novel conceptual ideas, which can be incorporated into both existing and future research. An in-depth hyperparameter tuning and evaluation of maximum performance against other state-of-the-art models and across a variety of use cases was considered out-of-scope, but remains a topic of future research.

\begin{figure*}
    \centering
    \includegraphics[width=1\textwidth]{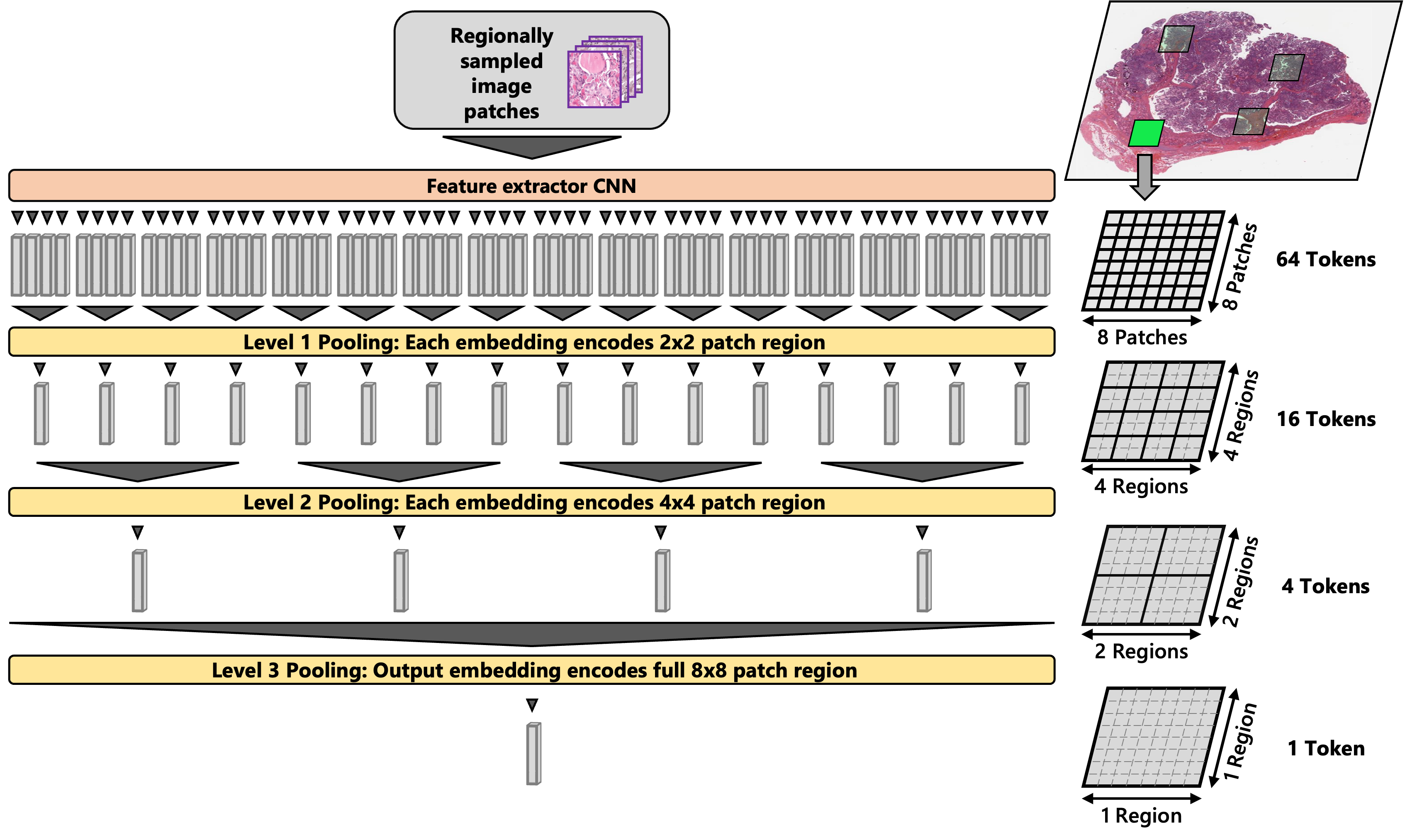}
    \caption{Illustration of embedding aggregation for a 3-level model with $2\times2$ regions. Each green square represents one $8\times8$ patch region. The depicted flow shows how one such region is reduced from 64 embeddings to a single aggregated representation. The size of the full region (green) depends on the number of aggregation levels (3) and the size of the aggregated regions ($2\times2$) on each level, which are configurable hyperparameters. The pooled embeddings on every level can be passed into the global ViT module (see Figure \ref{fig:architecture}).
    }
    \label{fig:multilevel}
\end{figure*}

\subsection{Transformer based aggregation}

For the Transformer model, we can formulate the attention mechanism as 
\begin{equation}
    \label{eqn:attention_general}
    \mathbf{Z} = Attention(Q, K, V) = softmax(QK^\top)V,
\end{equation}
where $Q$, $K$ and $V$ all represent the input bag $\hat{X}$ \cite{vaswani2017attention}. In practice, projecting $Q$, $K$ and $V$ linearly for $h$ multiple times for different attention heads has been shown to improve model performance \cite{vaswani2017attention}. The resulting aggregates for each head are concatenated and projected back to the original input dimension $d$. As described in Equation \ref{eqn:attention_weights}, the bag classification is based on a learnable parameter C with the same dimensionality $d$ as the input tokens in $\hat{X}$. The Vision Transformer (ViT) \cite{dosovitskiy2020image} uses such a class token for classification. Note that the self-attention mechanism described in Equation \ref{eqn:attention_general} is applied to all input tokens, and therefore outputs aggregations with regard to each token in the bag. It is therefore possible to stack multiple attention layers, as the number of input tokens stays consistent. The class token $C$ is only appended to the last layer in such cases, as it does not provide relevant information in previous layers. In our experiments, however, we only use one such attention layer and consequently only consider the $C$-based aggregated embedding for further processing in later stages.

The proposed model architecture is depicted in Figure \ref{fig:architecture}. Instead of feeding flattened image patches into the model as done in the standard ViT approach \cite{vaswani2017attention}, we first embed all sampled patches using a Resnet18 \cite{he2016deep} pretrained on ImageNet \cite{deng2009imagenet}. The fully connected layer at the end is removed to obtain 512-dimensional image-embeddings for each patch. For our experiments, we choose a ResNet18 for its relatively small memory footprint to enable online training of the patch-embedder. The architecture and pretraining of the patch embedder can be considered tunable hyperparameters.

 \subsection{Sampling}
Instead of randomly sampling patches globally, we first sample $N$ subregions from the input image. Patches are then sampled from from each subregion separately. Each subregion has the same size as determined by the model hyperparmeters. In a single level model, each subregion's patches are aggregated into a single embedding by the regional ViT using the learned class token $C$. All subregional embeddings are then passed into the global ViT to derive a slide-level representation. When using multiple levels, patches inside a subregion are pooled multiple times based on proximity. Figure \ref{fig:multilevel} depicts this process for a three-level model with $2\times2$ aggregation: first, all $2\times2$ regions are pooled into a single embedding. This process is repeated at the next level, aggregating $2\times2$ embeddings, each itself representing a $2\times2$ region, into a single embedding. These new embeddings, each representing a $4\times4$ patch region, are then pooled again on level 3 to derive a single embedding for the $8\times8$ subregion they belong to. At every level, the regionally pooled embeddings are also fed into the global ViT to derive a slide-level representation based on each aggregation level. This aims to allow the model to consider local interactions at different distances and creates an artificial "zoom-out" that processes features at different distances, thereby enabling it to learn morphological features larger than individual patches. 

We can derive the total side length ($TSL$) in number of patches of a sampling region depending on the side length $S$ of the aggregation window per level and the number of levels $L$ as shown in Equation \ref{eqn:base_size}:

 \begin{equation}
    \label{eqn:base_size}
    Total Side Length= S * 2^{L-1}.
\end{equation}

In consequence, increasing $S$ can drastically increase the total region size: in a 3-level model ($L=3$), a base region patch width of $S=2$ leads to a $TSL$ of 8 and therefore a region with 64 total patches. An increase of $S$ by 1 already increases the $TSL$ to 27, resulting in a region covering 729 patches. It is therefore clear that sampling on every level is required for higher values of $L$ and especially $S$. In future research, adjustable aggregation window sizes per level could also be explored.

\begin{table*}
    \centering
    \caption{Experiment AUC results. For all patches results, all non-empty patches from the input image were passed through the model. For top patches, only selected high attention patches were used as input as described in section \ref{sec:inference}.}
    \label{tab:Results}
    \begin{tabular}{l | c | c c |c c}

    &\textbf{Patch}&  \multicolumn{2}{c}{\textbf{CAMELYON16}} & \multicolumn{2}{|c}{\textbf{TCGA-THCA}}\\
    
    \textbf{Model}              & \textbf{selection} & \textbf{Mean} & \textbf{Best} & \textbf{Mean} & \textbf{Best}                    \\
    \hline
    \textbf{(1)} Baseline \cite{ilse2018attention}  & All & 0.764 $\pm$ 0.050 & 0.845 & 0.844 $\pm$ 0.010 & 0.855                       \\
                                                    & Top & 0.865 $\pm$ 0.020 & 0.886 & 0.853 $\pm$ 0.007 & 0.863                       \\
    \hline
    \textbf{Single level (ours)}                    &     &                   &       &                                                 \\
    \textbf{(3)} $2\times2$                         & All & 0.763 $\pm$ 0.035 & 0.833 & 0.852 $\pm$ 0.013 & 0.869                       \\
                                                    & Top & 0.898 $\pm$ 0.018 & 0.930 & 0.851 $\pm$ 0.015 & 0.870                       \\
    \hdashline
    \textbf{(4)} $3\times3$                         & All & 0.822 $\pm$ 0.039 & 0.873 & 0.855 $\pm$ 0.013 & 0.872                       \\
                                                    & Top & \textbf{0.914 $\pm$ 0.023} & \textbf{0.949}& 0.853 $\pm$ 0.025 & 0.878      \\
    \hline
    \textbf{Multi-level (ours)}                     &     &                   &       &                                                 \\
    
    \textbf{(5)} $2\times2-4\times4$                & All & 0.766 $\pm$ 0.055 & 0.864 & 0.882 $\pm$ 0.012 & 0.898                       \\
                                                    & Top & 0.867 $\pm$ 0.026 & 0.897 & 0.874 $\pm$ 0.019 & 0.897                       \\
    \hdashline
    \textbf{(6)} $2\times2-4\times4-8\times8$       & All & 0.735 $\pm$ 0.093 & 0.844 & \textbf{0.891 $\pm$ 0.014} & \textbf{0.905}     \\
                                                    & Top & 0.823 $\pm$ 0.042 & 0.892 & 0.888 $\pm$ 0.018 & 0.905                       \\
    \hdashline
    \textbf{(7)} $3\times3-9\times9$                & All & 0.777 $\pm$ 0.046 & 0.845 & 0.880 $\pm$ 0.010 & 0.893                       \\
                                                    & Top & 0.832 $\pm$ 0.030 & 0.863 & 0.875 $\pm$ 0.011 & 0.889                       \\
    \hdashline
    \textbf{(8)} $3\times3-9\times9-27\times27$~    & All & 0.799 $\pm$ 0.028 & 0.839 & 0.887 $\pm$ 0.017 & 0.905                       \\
                                                    & Top & 0.857 $\pm$ 0.041 & 0.899 & 0.884 $\pm$ 0.015 & 0.899                       \\
\end{tabular}
\end{table*} 

For our experiments, we use both $2\times2$ ($S=2$) and $3\times3$ ($S=3$) base patch regions. To enable comparability, we sample 4 patches per region in both approaches. We thereby can compare how the size of the region and the associated sparsity effects model performance. For the multi-level experiments, sampling is performed at all levels, i.e. in a 3-level $3\times3-9\times9-27\times27$ model, a total of 64 patches are selected from within the $27\times27$ region. We adjust the number of regions across all experiments such that there are always 256 patches sampled per whole slide image per epoch.

We use identical architectures for global and regional ViT modules, but only share weights between the regional modules, while the global ViT retains a separate set of weights. This is done to train the model to recognize similar structures at different scales. However, the individual configurations, including number of encoders, the size of the hidden dimension, the number of attention heads and the extent to which weights are shared between modules are adjustable hyperparameters and can be tuned further. Because the model architecture is inherently designed to aggregate patches sampled from the same region and thereby implicitly introduces information about regionality, we do not use positional embeddings. Exploring the use of relative positional embeddings for large region sizes could be subject to further research.

\begin{figure*}
    \centering
    \includegraphics[width=1\textwidth]{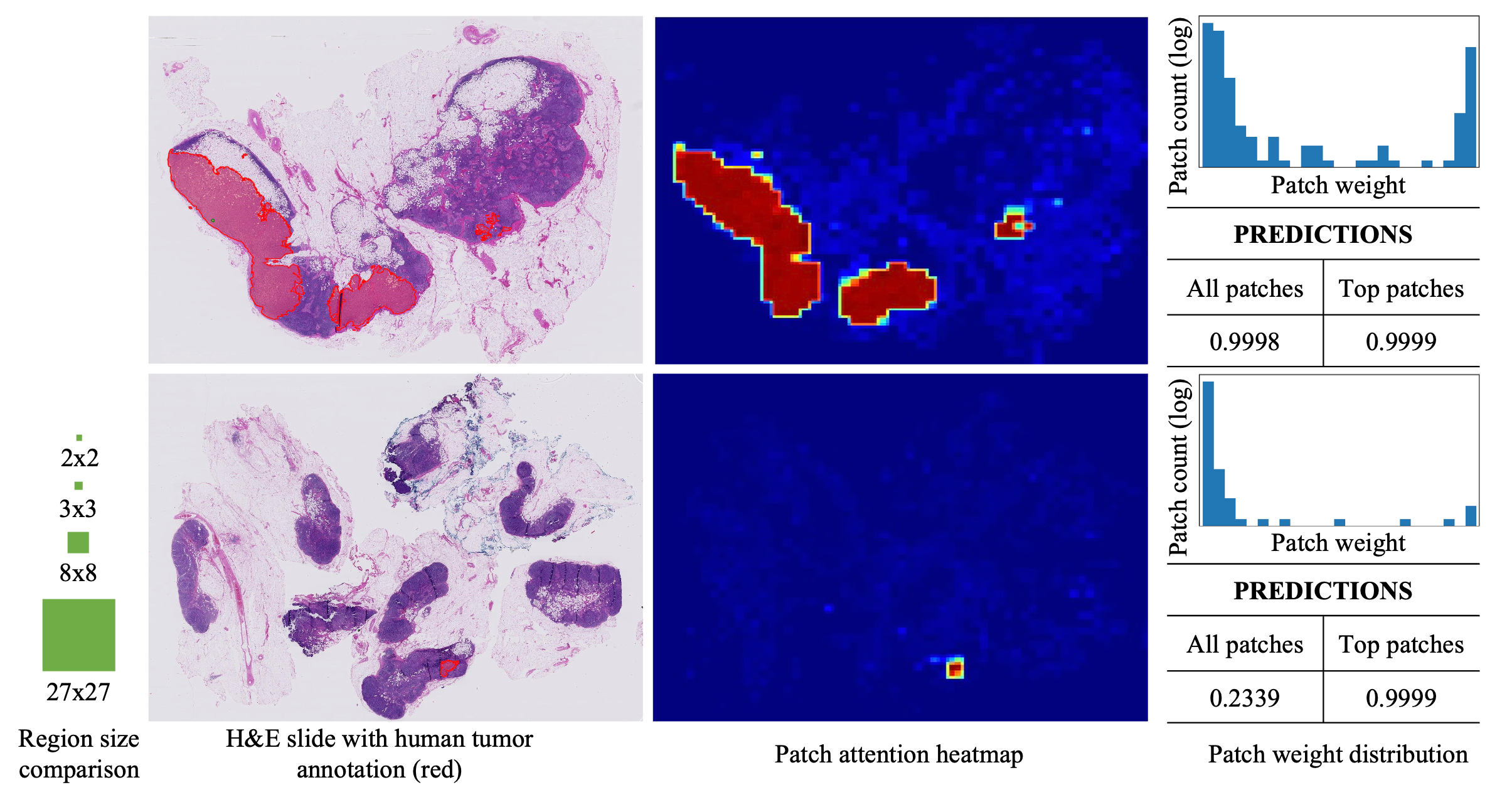}
    \caption{Comparison of two positive slides from CAMELYON16. As shown by the approximate visualizations of sampling regions, the high-attention regions of interest can be very small, extending only across few patches. Because larger aggregation regions contain a less representative sample of the full image, it is easy to see how small ROIs can be missed during sampling, hindering efficient model training. Conversely, the value of our high-attention inference method can be directly seen for cases where only few patches are relevant. This is also visible in the histograms of attention weight distributions. Both images show two peaks of high and low attention patches. However, the upper image has relatively larger tumor regions with many relevant patches and therefore receives a positive score with high confidence when predicting using all patches. For the lower image with a single small region with fewer positive patches, the prediction improves drastically when processing high attention patches.}
    \label{fig:slides}
\end{figure*}

\subsection{Inference}\label{sec:inference}
During inference, the complete slide is tiled into regions and the processed by the model. This is done by first computing an embedding of all patches and then passing the embeddings through the attention mechanism, thereby significantly reducing the memory footprint.

Additionally, after deriving attention weights for all patches in the first inference pass, we apply k-means clustering with two centroids to the attention values to flexibly select high-attention patches. This step can be considered as fitting a univariate gaussian mixture model to estimate two separate distributions of high-and low attention patches respectively. We then repeat inference with all patch embeddings belonging to the low attention cluster zeroed out. This method aims to reduce the effect of noise (i.e., irrelevant patches) on the prediction, especially in cases with small regions of interest, where even a strong signal from individual patches / regions can be diluted in the attention based aggregation process.

\subsection{Implementation Details}

Models are trained for 50 epochs using the AdamW optimizer with a learning rate of \textit{2e-5} and a batch size of 2. In all experiments, 256 patches are sampled per image per epoch. All regional sampling methods sample 4 patches per lowest level. We apply random augmentations on every patch including horizontal and vertical flips, sharpness and contrast adjusting and light blurring. All experiments are repeated 5-times using different seeds for sampling, but keeping the train-validation-test split consistent.

\subsection{Data}

We evaluate our model on two publicly available datasets containing WSI of H\&E-stained tissue samples. The CAMELYON16 dataset \cite{bejnordi2017diagnostic} depicts sentinel lymph nodes containing breast cancer metastases. We keep the train-test split predetermined by the dataset, resulting in 270 slides used for training and 129 slides used for testing. Of the train dataset, 85\% is used for training and 15\% for validation.

TCGA-THCA is a dataset of whole slide images of thyroid tissue provided by The Genome Cancer Atlas. The prediction of V600E mutation status of the BRAF gene mutations based on thyroid WSI has been subject of recent research \cite{anand2021weakly,hohne2021detecting}. It contains a total of 482 tissue slides, of which 294 are labeled as BRAF positive. Of the data, 65\%, 15\% and 20\% of cases are used for training, validation and testing respectively.

\subsection{Preprocessing}
 All images are tiled into patches with an edge size of 157 micrometers, which are then resized to a size of $224\times224$ pixels. Empty patches are removed using threshold-based filtering. Luminosity standardization followed by stain normalization \cite{vahadane2016structure} is then applied to the patches to reduce variance between images caused by differences in staining procedures and equipment.

\section{Results}

\begin{figure*}
    \centering
    \includegraphics[width=1\textwidth]{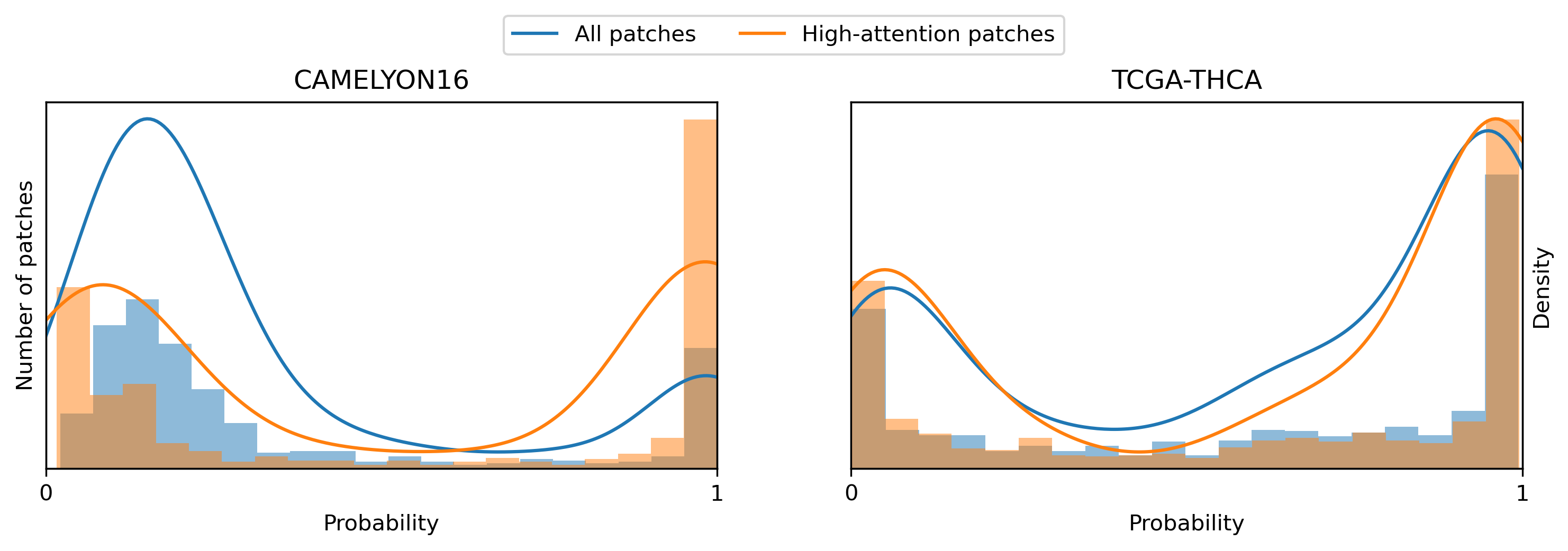}
    \caption{Predicted probabilities per slide, averaged across all seeds, using all patches and our high-attention inference strategy. Results are based on the respective best performing model (see Table \ref{tab:Results}). The high-attention strategy moves predictions closer to the extremes of 0 and 1. This is especially true for CAMELYON16. For TCGA-THCA, the effect is less pronounced, which is likely to the much larger amount of high attention patches in most images.}
    \label{fig:distribution}
\end{figure*}

The results of our experiments are shown in Table \ref{tab:Results}. On the CAMELYON16 dataset, we observe a notable performance increase over the baseline for both single-level regional aggregation models. Especially the $3\times3$ model achieves the highest average AUC of 0.914 as well as the highest AUC observed overall of 0.949. We assume that the $3\times3$ region size is well suited for the morphological size of tumors in the dataset. Additionally, the sparsity of sampling in the $3\times3$ regions, compared to $2\times2$, might allow the model to better handle patches which are not directly adjacent. In contrast, multi-level approaches yield a slightly reduced AUC when increasing the total aggregated region size, while also having larger standard deviation in results. We hypothesize the better performance of regional, Transformer based aggregation on small regions is driven mainly by the small size of regions of interest in many of the images in the dataset. As most images contain tumors spanning across only few patches, there is little potential benefit by aggregating larger regions.

On TCGA-THCA, the single level approaches do not perform notably better than the baseline. However, we see a significant improvement of results when sampling and aggregating larger regions in multiple steps. The three-level approaches in both cases outperform the respective two-level approaches. As the total region size of $9\times9$ for experiment (7) is larger than that of (6), we theorize that the additional aggregation layer does help to improve predictive performance by exploiting the joint information content across a larger region. The results point towards the increased ability of the multi-level model to leverage the comparatively large tumor regions in the TCGA-THCA dataset. Furthermore, the effectiveness of 3-level approaches over 2-level approaches suggests there is potential in deriving image representations at different "zoom" levels.

For all our experiments, we report results on the full image as well as for running inference only on high attention patches. We see significant differences in the effectiveness of this approach between the two datasets. For CAMELYON16, we find a notable improvement in AUC both for the baseline as well as for all of our approaches. As discussed above, this approach is likely particularly effective for CAMELYON16 because the signal of very small regions of interest gets diluted by large amounts of noise (i.e., patches with no relevant information about absence / presence of tumorous tissue). An example of this can be seen in Figure \ref{fig:slides}, where the prediction of the slide with a small tumor area changes drastically. Conversely, we observe no positive effect of top patch inference on TCGA-THCA, where tumor regions are much larger on average. The results highlight the potential of the proposed inference strategy, but also demonstrate that the expected gains are highly dependent on the specific morphological challenges present in any given dataset.

We were also interested whether selecting only high-attention patches would move all predictions towards positive values. As shown in Figure \ref{fig:distribution}, this is not the case. The high-attention inference strategy does help to improve model certainty, both for positive and negative predictions. This indicates that the models are able to learn negative evidence which indicate the absence of tumorous tissue. More importantly, it therefore does show that performing inference on high-attention patches in negative slides does not cause the model to raise the assigned probability, underlining the robustness of our method.

\section{Conclusion}

In this work, we have demonstrated a method of regionally fusing patch embeddings to derive slide-level predictions. We show the importance of fitting the approach to the target dataset: For those with small ROIs, choosing a small, highly local aggregation method leads to best results. Datasets with large regions of interest containing complex morphological structures can be handled effectively by aggregating large regions in multi-step approaches. For both datasets, we are able to apply our method to effectively improve model performance compared to the baseline. In this work, we also suggest a simple, easily transferable method of performing inference on selected high attention patches that can significantly improve predictive results at little additional cost. Our results point towards several promising directions for further research, including an investigation into the effects of regional and global sampling with different region sizes as well as the analysis of the effects of noise on prediction results. Finally, future research could explore hyperparameter tuning and absolute performance evaluation against common state-of-the-art approaches \cite{lu2021data,shao2021transmil,li2021dual,chen2022scaling}, as well as expanding these methods to use a regional and/or hierarchical aggregation approach. Similarly, we consider the application of our method to other datasets with different characteristics as well as on multi-class problems a highly interesting topic.\\

\noindent\textbf{Acknowledgment.} The results shown here are in whole or part based upon data generated by the TCGA Research Network: https://www.cancer.gov/tcga.

{\small
\bibliographystyle{ieee_fullname}
\bibliography{egbib}
}

\end{document}